\documentclass[
  journal=large,
  manuscript=Position~Paper
]{cup-journal}

\usepackage{amsmath}
\usepackage[nopatch]{microtype}
\usepackage{booktabs}
\usepackage{cite}
\usepackage{hyperref}

\title{Dynamic semantic networks for exploration of creative thinking}

\author{Danko D. Georgiev}
\affiliation{Institute for Advanced Study, 30 Vasilaki Papadopulu Str., Varna, 9010, Bulgaria}

\author{Georgi V. Georgiev}
\affiliation{Center for Ubiquitous Computing, Faculty of Information Technology and Electrical Engineering, University of Oulu, Oulu, FIN-90014, Finland}
\email[Georgi V. Georgiev]{georgi.georgiev@oulu.fi}


\begin{document}

\begin{abstract}
Human creativity originates from brain cortical networks that are specialized in idea generation, processing, and evaluation. The concurrent verbalization of our inner thoughts during the execution of a design task enables the use of dynamic semantic networks as a tool for investigating, evaluating, and monitoring creative thought. The primary advantage of using lexical databases such as WordNet for reproducible information-theoretic quantification of convergence or divergence of design ideas in creative problem solving is the simultaneous handling of both words and meanings, which enables interpretation of the constructed dynamic semantic networks in terms of underlying functionally active brain cortical regions involved in concept comprehension and production. In this study, the quantitative dynamics of semantic measures computed with a moving time window is investigated empirically in the DTRS10 dataset with design review conversations and detected divergent thinking is shown to predict success of design ideas. Thus, dynamic semantic networks present an opportunity for real-time computer-assisted detection of critical events during creative problem solving, with the goal of employing this knowledge to artificially augment human creativity. 
\end{abstract}

\section{Introduction}


Creativity is the capacity to use imagination and inventiveness to bring into existence original ideas, solutions or products.
Ideas and products are judged as creative to the extent that they provide both an original and valuable solution to the problem at hand \cite{Srinivasan2010,Casakin2021,Moldovan2011}, and the problem is heuristic rather than algorithmic \cite{Amabile1983a,Amabile1983b}.
If an idea is not new but already known, then we could just algorithmically copy the known idea investing no creative efforts.
Alternatively, if a new solution of a posed problem is not useful, we would consider bringing it into existence as a waste of available resources.
Therefore, originality (novelty) and usefulness (value) of ideas interweave
together as parts of a single creativity construct \cite{Stein1953,Wang2013,Lee2020,Taura2016}.


The most creative ideas are both novel and useful, and the most creative people excel at both creativity dimensions.
Descriptive psychological accounts of creativity \cite{Guilford1957,Hudson1974,Runco2004,Runco2007,Runco2020}
postulate that novelty is generated through divergent (associative, analogical) thinking that breaks assumptions and rules,
while usefulness is enhanced through convergent (analytical) thinking that adheres to needs, boundaries, and constraints \cite{MironSpektor2017}.
Both convergent or divergent paths are instructed in the design process \cite{Goldschmidt2016,Goldschmidt2019}.
The process of innovative abduction is also related to both divergent and convergent thinking \cite{Dong2016}.
Such descriptive accounts could be implemented in artificial intelligence machines only if translated into quantitatively precise procedures.


Cognitive processes and structures that underpin creative thinking and help produce creative acts and results are referred to as creative cognition \cite{Finke1992}.
While the dynamic cognitive processes in creative contexts have been in the focus of research \cite{Wilkenfeld2001,Sonalkar2016}, objective tools for evaluation of creative cognition have only recently been developed  \cite{Gogo2018,Gogo2019,Gogo2019b,Han2020,Han2021,Han2022,Chiu2023}.
For example, dynamic semantic networks have been used to analyse
the datasets from the 10th~Design Thinking Research Symposium (DTRS~10) with design review conversations between design students and real clients recorded in educational settings \cite{Adams2013,Adams2015,Gogo2018}
and from the 12th~Design Thinking Research Symposium (DTRS~12) with design discussions in a company context \cite{Christiaans2018,Gogo2023}.
Several semantic measures were found to be useful for quantitative evaluation of convergence/divergence in creative thinking and supported the role of divergent thinking for the success of generated design ideas \cite{Gogo2018,Gogo2023}.
{Procedures for real-time application of dynamic semantic networks in creative problem solving, however, are currently lacking. This necessitates the development of a detailed workflow for real-time application of dynamic semantic networks for monitoring and potential support of design creativity.}

\subsection{Main hypothesis and aims of the study}

{The main hypothesis of this work is that dynamic semantic networks could be used in real time to predict the success of creative design ideas.
To test this hypothesis, first we aimed to develop a complete workflow for real-time application of dynamic semantic networks, using a moving time window for monitoring the cognitive processes during creative design.
Second, we aimed to identify particular quantitative measures of information content and semantic similarity that highly correlate with human evaluation of word similarity in order to employ those measures in the constructed dynamic semantic networks for modeling creative cognition.
Third, we aimed to evaluate the actual performance of the developed workflow through back-testing the dynamic semantic networks for predicting the success of design ideas using transcribed design review conversations from the DTRS~10 dataset.}

\section{Aim 1: Workflow for monitoring of creative cognition}

{The inner privacy of consciousness poses unique challenges to understanding cognitive processes \cite{Georgiev2017,Georgiev2020,Georgiev2020b}.
Experiences, feelings, emotions, thoughts and beliefs constitute one's consciousness, but the phenomenological qualia of these conscious states are not directly accessible by external observers \cite{Nagel1974}. Instead, the individual conscious states need to be externalized through expression into classical bits of Shannon information \cite{Shannon1948} such as words, gestures or images. Because verbalized reports of the individual stream of consciousness concurrent with the execution of a given cognitive task can be used reliably as data \cite{Ericsson1980} for subsequent analysis with natural language processing scripts \cite{Bird2009}, we have developed a workflow for monitoring of creative cognition based on design conversations that occur during the development of a particular design product.}

\begin{figure}[t!]
\begin{centering}
\includegraphics[width=0.7\textwidth]{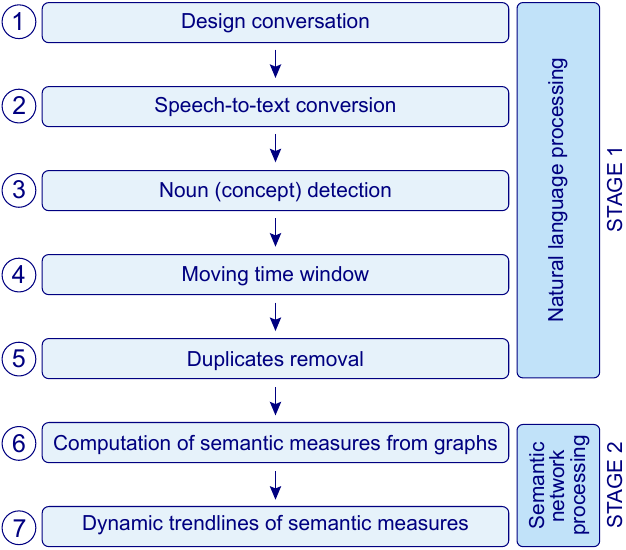}
\par\end{centering}
\caption{\label{fig:1}Workflow for monitoring of creative cognition with dynamic semantic networks.}
\end{figure}

{The overall structure of the workflow consists of 2 stages: natural language processing and semantic network processing (Fig.~\ref{fig:1}). The first stage consists of 5 steps, all of which can be automated with the use of available Python libraries and scripts: (1) audio recording of the design conversation, (2) speech-to-text conversion, (3) part-of-speech tagging for detection of nouns (concepts), (4) construction of a moving time window and (5) removal of duplicates. The second stage consists of 2 steps: (6) construction of semantic networks based on WordNet, including computation of quantitative semantic measures from graphs, and (7) dynamic statistical fit of trendlines for the selected semantic measures for detection of convergence or divergence in creative thinking.}

{In this work, we consider that the engineering solutions for natural language processing needed to accomplish stage 1 of the workflow are readily available \cite{Bird2009,Loria2016,Lee2024}. Therefore, we will dedicate our efforts to providing detailed procedures for executing the semantic network processing using WordNet in stage 2 of the workflow. We will also elaborate on different graph theoretical alternatives for computation of \emph{information content}, which is a measure of the surprisal due to the occurrence of a particular concept, and \emph{semantic similarity}, which is a measure of how close the meanings of two concepts are. The dynamic trendlines for information content and semantic similarity obtained from recorded design conversations will then be back-tested for correlation with the eventual success of developed design products using the DTRS10 dataset and we will propose potential application of dynamic semantic networks for real-time support or enhancement of design creativity.}

\section{Aim 2: Modeling creative cognition with dynamic semantic networks}

\subsection{Replacement of subjective expert coding with objective computer algorithms}

Creativity could be modeled computationally \cite{Sosa2021}.
{The theoretical foundations of automated problem solving are based on automated reasoning systems based on heuristic search techniques such as General Problem Solver \cite{Newell1959}. Early theories of problem-solving use computer simulations to predict human performance, explain the underlying processes and mechanisms, account for incidental phenomena, show how performance changes under different conditions, and explain how problem-solving skills are learned \cite{Simon1971}. The challenges of computational problem-solving in design revolve around encoding, representation, constraint analysis, and reduction of the effective solution space \cite{Perkowski2022}.}

Protocol analysis of concurrent verbalization from professional design teams allows for the identification of reasoning patterns in idea generation \cite{Cramer2019},
dynamic process patterns in design communication \cite{Cash2020},
success of design ideation \cite{Maccioni2020} or
evaluation of external sources of inspiration \cite{Borgianni2020}.
Typically, the protocol analysis requires protocol coding performed by an expert, which invariably introduces a level of subjectivity that limits reproducibility by independent research teams. 
{Furthermore, significant attention has been focused on the subjective nature of evaluating creativity of design ideas using metrics \cite{Fiorineschi2023}. The subjective judgments required in metrics can significantly impact the evaluation of design creativity, owing to differences in how evaluators perceive and prioritize different aspects of the design ideas \cite{Fiorineschi2022}. Hence, further research is needed to address these challenges and develop more robust methods for evaluating design ideas
\cite{Borgianni2020}.}
To dispense with the reliance on experts and ensure maximal reproducibility, here we describe an objective method for textual processing and construction of semantic networks that is fully automated by applied computer algorithms.

\subsection{Semantic networks as graphs}

\begin{figure}[t!]
\begin{centering}
\includegraphics[width=\textwidth]{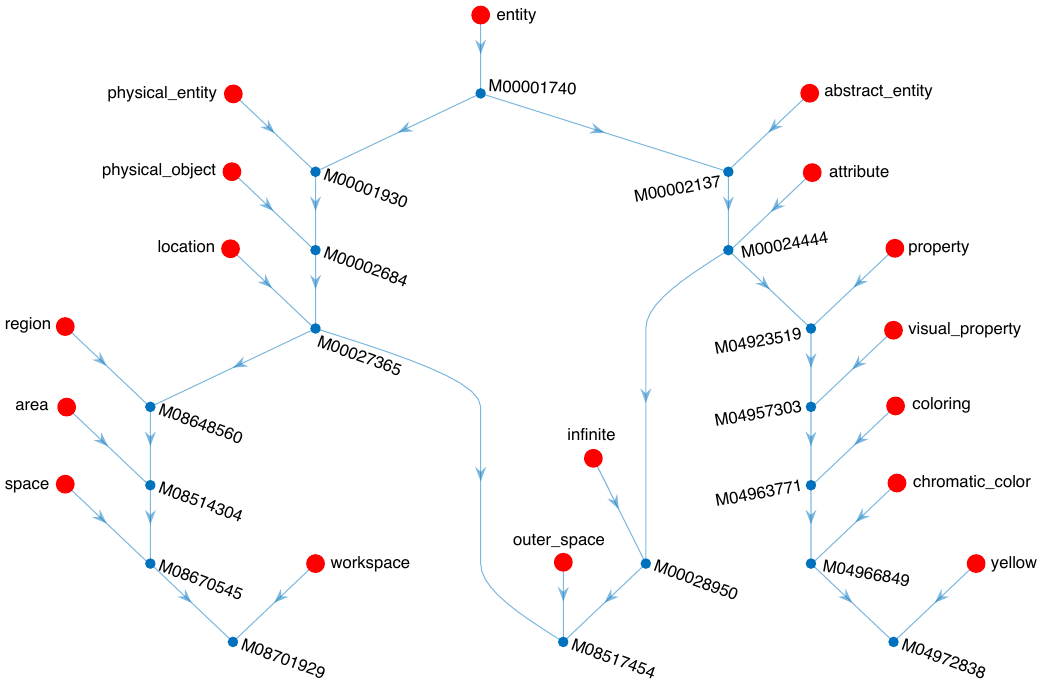}
\par\end{centering}
\caption{\label{fig:2}The subgraph of meanings $M$ in WordNet~3.1 does not have directed cycles when the edges remain directed, however, when all edges are converted into undirected edges using the graph operator $U$, the graph $U(M)$ becomes cyclic. One consequence of the structure of $M$ is that even for two monosemous words such as `workspace' and `yellow' the shortest path in the undirected graph $U(M)$ may not pass through the lowest common subsumer, which in this case is the root meaning vertex $M00001740$. In this example, the length of the shortest path between `workspace' and `yellow' is 12, whereas the distance between `workspace' and `yellow' through their lowest common subsumer $M00001740$ is 14. To avoid clutter in the image, we have added only a single word for each meaning vertex, however, many of the meaning vertices are subsumed by synsets of words. For example, both words `yellow' and `yellowness' subsume the meaning vertex $M04972838$.}
\end{figure}

Semantic networks represent knowledge in the form of \emph{graphs} that consist of \emph{vertices} and \emph{edges} \cite{Diestel2017}. Vertices indicate individual concepts, whereas edges between pairs of vertices indicate specific semantic connections \cite{Boden2004}.
Graphs could be divided into directed or undirected depending on the type of edges contained in the graph. 
The lack of loops, which are edges that connect vertices to themselves, and multiple edges with the same source and target vertices generates a simple directed graph.
The lack of directed cycles generates a directed acyclic graph (DAG).
The underlying undirected graph of a DAG, however, could contain cycles
(Fig.~\ref{fig:2}).
Further introduction of a root vertex such that all edges of the directed graph are directed either away from or towards the root generates a rooted directed acyclic graph.
Assigning different weights to the edges of directed or undirected graphs generates weighted graphs.

WordNet is a lexical database, which represents human knowledge in a graph form \cite{Miller1990,Miller1995,Miller1998,Fellbaum1998a,Fellbaum1998b}. This is particularly suitable for imposing a distance function onto constructed semantic networks, which in turn enables quantitative exploration of dynamic cognitive processes such as creative cognition.
Throughout this work, we employ WordNet~3.1, which is represented by a \emph{rooted directed acyclic graph} of meanings (Fig.~\ref{fig:2}).
The constructed semantic networks are represented by \emph{undirected
cyclic weighted graphs} (Fig.~\ref{fig:3}).

\subsection{Semantic networks as models of the creative mind}

Semantic networks constructed from conversation transcripts
represent computational models of conceptual associations and structures
\cite{Gogo2008,Gogo2010,Han2020,Han2021,Han2022}. Individual concepts in the semantic network need not be only single words, but could also be phrases combining different parts of speech \cite{Esparza2019}, and can employ specific technical concepts \cite{Sarica2021}.
However, for ensuring reproducibility and objectiveness, the extraction of concepts for the semantic networks should be based on a set of rules that can be automated by a computer program without the need of human intervention.
This problem is addressed satisfactorily by identifying individual
concepts with single words that could be further narrowed down to
a single lexical category (nouns) with the use of part-of-speech tagging
performed by natural language processing software \cite{Gogo2018}.

\subsection{Polysemy necessitates simultaneous use of words and meanings}

Polysemy is essential for creative conceptual blending and associative thinking \cite{Nerlich2003,Fauconnier2003,Gogo2014,Ravin2000}. In particular, designer's mind could work simultaneously with several meanings of a word, similarly to how writers employ polysemy in humorous works \cite{Boxman2014}. Meaning is deemed an essential component of creativity \cite{Saaksjarvi2018}.
In order to capture the possible impact of polysemous words in design thinking, we employ a computational method that does not compromise, reduce, or disambiguate between the multiple senses.
The linguistic distinction between \emph{words} and \emph{meanings} could be approached in two different ways. Inclusion of a pre-processing step called \emph{disambiguation of senses} could convert all words into senses. This would modify the verbal data through injection of interpretation
even before the data analysis is started and would delete potentially
useful information about underlying difficult-to-observe cognitive
processes. For example, the nouns entering into the description of
creative ideas may acquire different senses that disagree with those
listed in a dictionary \cite{Gogo2014,Taura2013}. Furthermore, the
polysemy of nouns was found to be instrumental in the association
of ideas, which were previously not considered to be related by the
creative problem solver \cite{Gogo2014}.
To avoid the latter shortcomings, we construct semantic networks from verbal data without any disambiguation of senses of extracted words (nouns) \cite{Gogo2018}, but preserving two types of vertices in the semantic network, word vertices and meaning vertices. The benefit is that discovered functional relationships in the semantic network could be correlated to the neural activity in specialized brain cortical areas such as Broca's area, which translates meanings into words, and Wernicke's area, which translates words into meanings \cite{Georgiev2021}.
The utility of semantic networks of nouns for studying creativity and reconstruction of difficult-to-observe cognitive processes in conceptual design was demonstrated previously by different research teams with the use of several experimental datasets \cite{Gogo2008,Gogo2010,Taura2012,Yamamoto2009}.

\subsection{Creative cognition modeled as a dynamic semantic network}

\begin{figure}[t!]
\begin{centering}
\includegraphics[width=\textwidth]{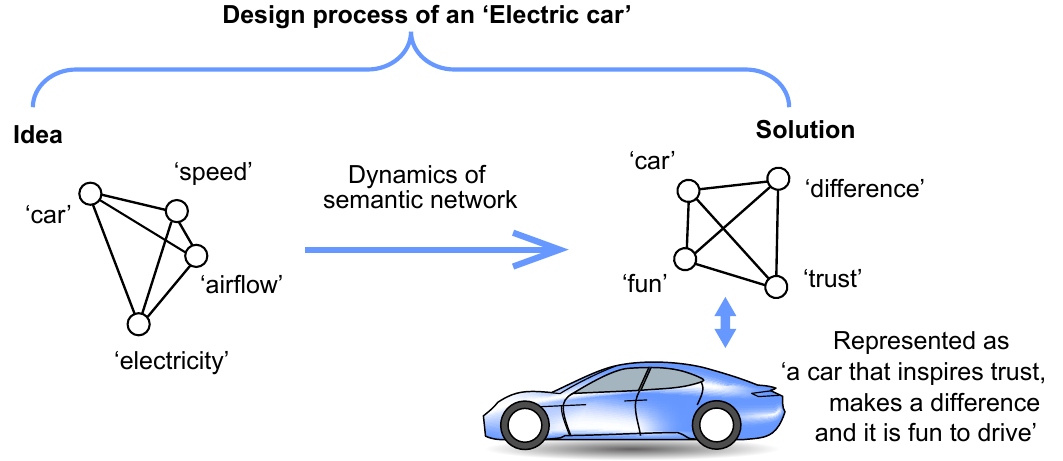}
\par\end{centering}
\caption{\label{fig:3}Modeling the creative design process of an `Electric car' with a dynamic semantic network. The dynamics of the semantic network from the stage of idea generation to the stage of fully developed solution provides a useful, reproducible and fast computational tool for exploration of creative thinking.}
\end{figure}

To assess the dynamic aspects of cognitive processes, the constructed
semantic network should be allowed to evolve in time (Fig.~\ref{fig:3}).
One inefficient approach is to consider the cumulative growth of the
semantic network as new concepts appearing in the transcribed conversations
are added to those concepts that have already appeared.
This cumulative approach creates large networks fast even for relatively short conversations, which produces a large sample size for statistical analyses.
The disadvantage is that it retains in the semantic network concepts
that may have been briefly considered but then discarded by the cognitive
processes underlying the given task.
An alternative, much more efficient approach is to coarse-grain the time into time intervals, each of which will encompass a corresponding part of the conversation \cite{Gogo2018}.
The advantage of the noncumulative approach is that concepts are retained in the semantic network only if they repeatedly appear in the course of the problem solving conversation.
The minimal time interval should be coarse-grained in order to contain a sufficient number of concepts to form a meaningful network.
For real-time monitoring of verbal output, the semantic network dynamics could be tracked with a \emph{moving time window} allowing for dynamic update of the semantic measures with each new word added in the conversation.

\section{WordNet structure as a graph of meanings and words}

\subsection{Lexical categories in WordNet}

WordNet~3.1  is publicly available (http://wordnet.princeton.edu), lexical database for English created under the direction of G. A. Miller and hosted at Princeton University \cite{Fellbaum1998a,Fellbaum1998b,Miller1990,Miller1995}.
WordNet contains four subnets that correspond to four basic lexical
categories: nouns, verbs, adjectives, and adverbs \cite{Fellbaum1998b,Miller1998}.
Because there are only a few cross-subnet pointers \cite{Fellbaum1998b},
calculation of graph-theoretic distances between words could be achieved
by confining the constructed semantic networks to a single subnet.
The subnet of nouns provides the largest and deepest hierarchical
taxonomy in WordNet, which can be efficiently used for the construction
of semantic networks of nouns. In an experimental study using design
review conversations, it was found that over 99.8\% of all nouns
used in the conversations are also listed in WordNet~3.1 \cite{Gogo2018}.
Further motivation for working with nouns in research of creative
problem solving is provided by developmental linguistics findings
that the category corresponding to nouns is, at its core, conceptually
simpler or more basic than those corresponding to verbs and other
parts of speech, which is exemplified by infants early advantage for
learning nouns over verbs \cite{Gentner1982,Waxman2013}. In addition,
experimental creativity research has shown that networks of nouns stimulate the generation of creative ideas \cite{Gogo2019}, different
combinations of nouns and relations between nouns are associated with
a display of creative thought \cite{Dong2009}, and dissimilarity
of noun pairs produces a higher number of emergent features of creative
ideas \cite{Wilkenfeld2001}.

\subsection{Hypernym-hyponym hierarchy of nouns}

The two basic semantic relations in WordNet are synonymy, where sets
of word synonyms (synsets) form the basic building blocks of the lexical
hierarchy, and hyponymy (subordination of synsets) where if a hyponym
(subordinate) $X$ is subsumed by a hypernym (superordinate) $Y$
then it follows that `An $X$ is a kind of $Y$' \cite{Miller1990,Miller1998}.
The hypernym-hyponym (is-a) relationship provides a taxonomy of nouns in WordNet as follows: the root synset \{entity\} subsumes directly three different synsets, which can be viewed as classifying entities into \{abstract\_entity, abstraction\}, \{thing\} or \{physical\_entity\}.
Each of these synsets then subsumes directly other synsets, further classifying classes of entities into subclasses, and so on.
The hypernym-hyponym (is-a) hierarchy of nouns in WordNet is
conceptually clear if it is represented in the form of a graph with
two types of vertices: \emph{meaning} vertices and \emph{word} vertices.

The distinction between \emph{words} and \emph{meanings} would not have been required if there were one-to-one relationship between words and meanings.
In natural language, however, one word can have several meanings (polysemy) and one meaning can be expressed by several words.
For example, the word `horse' has 5~meanings thereby entering into 5~synsets as follows: M02377103 with synset \{Equus\_caballus, horse\}, M03543217 with synset \{gymnastic\_horse, horse\}, M03629976 with synset \{horse, knight\}, M04147696 with synset \{buck, horse, sawbuck, sawhorse\} or M08414813 with synset \{cavalry, horse, horse\_cavalry\}. Here, `M' stands for meaning, the subsequent digits indicate the number by which the particular synset is referred to in WordNet. Explicitly labelling the meaning vertices with a numeric code further clarifies the significance of the synset and avoids possible misunderstanding of the synset as a \emph{list of words}---in fact, the synset stands only for the \emph{meaning that is in common for all words in the list}.
For example, the meaning of M02377103 with synset \{Equus\_caballus, horse\} is `the particular animal species in the genus Equus', whereas the meaning of M03543217 with synset \{gymnastic\_horse, horse\} is `an artistic gymnastics apparatus'. The fact that the \emph{meaning} is the essential ingredient of a synset can be illustrated with a somewhat rare example of a meaning, which is difficult to guess from the words in the synset alone: the meaning of M03629976 with synset \{horse, knight\} is actually `a chess piece shaped to resemble the head of a horse'---this has to be understood from a sample sentence used in WordNet to clarify the meaning.

\subsection{Composition of word and meaning subgraphs}

The lexical hierarchy of nouns in WordNet~3.1 is comprised of 158441
word vertices and 82192 meaning vertices. Word vertices
and meaning vertices are organized in two subgraphs, subgraph $M$,
composed of 84505 meaning$\to$meaning edges between hypernyms and
hyponyms, and subgraph $W$, composed of 189555 word$\to$meaning
edges. The subgraph $M$ is a rooted directed acyclic graph, which
has as a root the meaning vertex M00001740 corresponding to the single-word synset \{entity\}.
To compute graph theoretic measures for words,
however, the subgraph $M$ should be expanded with edges extracted from
the subgraph $W$. For example, if we are interested in any semantic
measure characterizing the word `horse', we will need to extract the
5 edges from $W$ that connect `horse' to each of its 5 meaning vertices.
Only in the composite graph containing both meanings and words, we are able to comprehend the content of transcribed conversation.

\begin{figure}[t!]
\begin{centering}
\includegraphics[width=\textwidth]{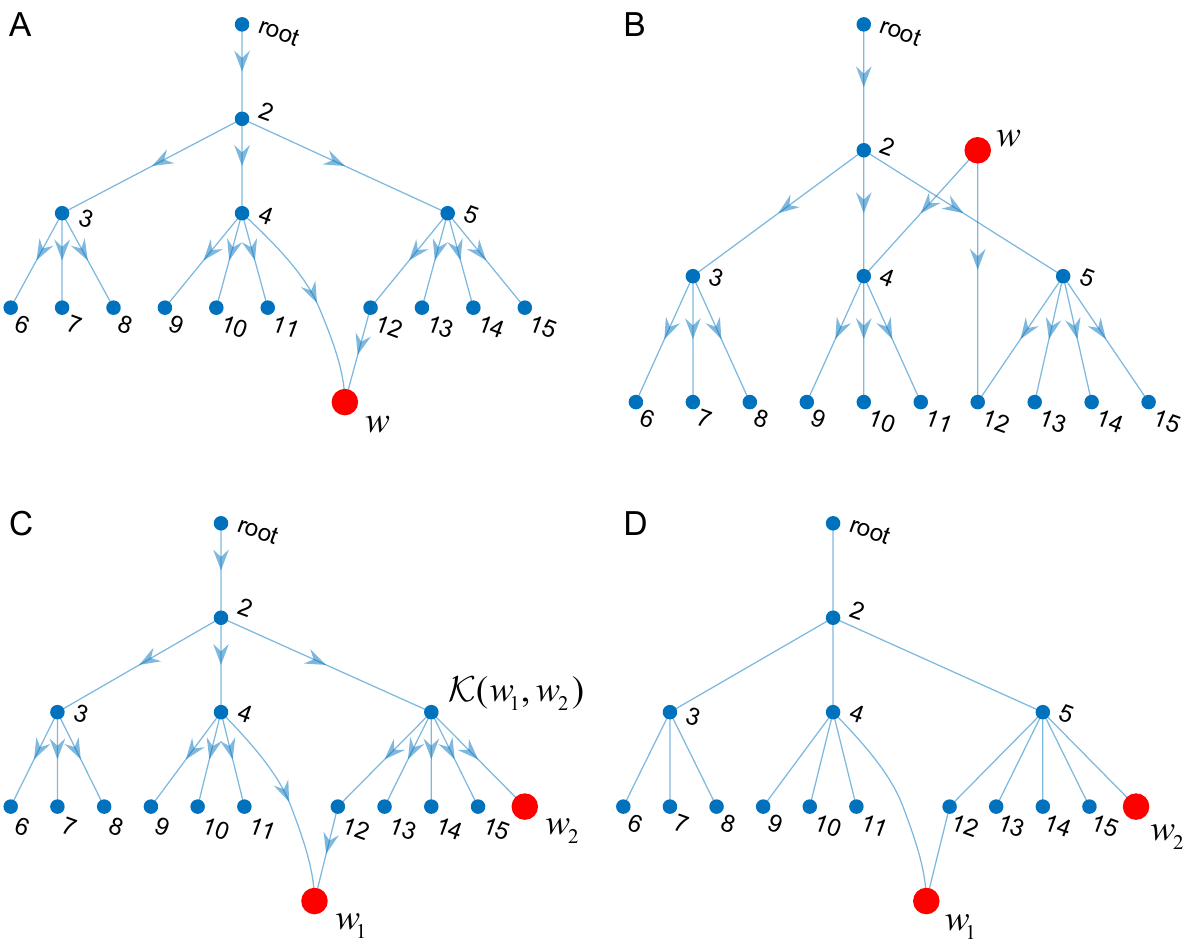}
\par\end{centering}
\caption{\label{fig:4}Graph composition of meaning vertices and word vertices
for different WordNet~3.1 searches. (A)~Adding words as subordinate
vertices subsumed by meanings allows for computing the depth in the
taxonomy and listing the meaning subsumers of words. (B)~Adding words
as subsumers of meanings allows for listing the meaning subvertices
and leaves. (C)~Adding two distinct words $\left\{ w_{1},w_{2}\right\} $
as subordinate vertices subsumed by meanings allows for computing
their lowest common subsumer $\mathcal{K}\left(w_{1},w_{2}\right)$.
(D)~Adding two distinct words $\left\{ w_{1},w_{2}\right\} $ into
an undirected graph allows for computing the shortest path distance
between the two words, which may not pass through their lowest common subsumer. Different graph compositions are needed for path-based or information content-based quantification of word similarity.}
\end{figure}

The reason for appending extracted word edges to the graph $M$, instead
of working in the full graph $M\cup W$, is to achieve computational
efficiency, namely, the subgraph $W$, which is effectively discarded,
has twice as many edges as the subgraph $M$ and finding shortest
paths is much faster in smaller graphs. Additional flexibility is
achieved by the possible application of directed graph operators such
as $R(G)$, whose action on the graph $G$ is to reverse the direction
of all edges, or $U(G)$, whose action on the graph $G$ is to remove
the directionality of all edges \cite{Gogo2018}. As an example, $W$ contains word$\to$meaning edges, $R(W)$ contains meaning$\to$word edges, and $U(W)$ contains meaning---word edges. In this way, it is possible to add words as
subordinate vertices (subsumed by meanings) or as superordinate vertices
(subsumers of meanings) depending on the semantic measures that need
to be computed (Fig.~\ref{fig:4}).

\section{Semantic measures based on WordNet}
\subsection{Graph-theoretic functions}

Defining semantic measures for words from the graph structure of WordNet requires the introduction of several basic functions that take as arguments a graph, denoted with capital letter, and one or more vertices, denoted with small letters \cite{Gogo2018,Gogo2023}.
Hereafter, general type of vertices, either meanings or words, will be denoted as $v_{1},v_{2},\ldots,v_{n}$, whereas word vertices will be specified as $w_{1},w_{2},\ldots,w_{n}$. 

\subsubsection{Functions in a general graph}

$\mathcal{J}(G,v)$ lists all edges that are incident to vertex~$v$ in the graph~$G$.

\noindent $\mathcal{A}(G,v)$ lists all vertices that are adjacent to vertex~$v$ in the graph~$G$.

\subsubsection{Functions in a directed graph}

$\mathcal{V}\left(G,v\right)$ lists all subvertices of vertex~$v$ in the graph~$G$, where subvertices are all vertices with finite directed path from $v$.

\noindent $\mathcal{S}(G,v)$ lists all subsumers of vertex~$v$ in the graph~$G$, where subsumers are all vertices with finite directed path to $v$.

\noindent $\mathcal{L}(G,v)$ lists the leaves of a vertex~$v$ in the graph~$G$, where leaves are all subvertices of $v$ with a vertex out-degree of zero.

\noindent $\mathcal{D}(G,v_{1},v_{2})$ gives the shortest path distance measured in edges from vertex~$v_{1}$ to vertex~$v_{2}$ in the graph~$G$, where the output is infinite $\infty$ if there is no path from $v_{1}$ to $v_{2}$.

\subsubsection{Functions in a rooted directed graph}

$\mathcal{T}(G,v)$ gives the depth in the taxonomy of a vertex~$v$ measured as the number of vertices on the shortest path from the root vertex $r$ to $v$ in the graph~$G$.

\subsubsection{Set theoretic functions}

$|f(x)|$ counts the number of elements in the list~$f(x)$.

\subsection{Semantic measures for single words}

The level of abstraction, polysemy and information content are semantic measures applicable to single words. For a semantic network composed of $n$ word vertices, the average for each of these semantic measures could be determined with $n$ searches in WordNet.

\subsubsection{Level of abstraction}

The level of abstraction of a word is the complement to unity of the
word concreteness. For nouns, both measures are related
to the noun depth in the WordNet taxonomy such that nouns located higher in
the hierarchy are more abstract and less concrete, whereas nouns located
lower in the hierarchy are less abstract and more concrete \cite{Gogo2018,Meng2013}.
In graph theoretic notation, the level of abstraction of the word~$w$ is
\begin{equation}
\text{Abstraction}(w)=1-\frac{\mathcal{T}(w)-1}{\mathcal{T}_{\max}-1}
\end{equation}
where $\mathcal{T}_{\max}=19$ is the maximal depth of WordNet~3.1 taxonomy and $\mathcal{T}(w)$ is the depth of the word~$w$ computed as the shortest path distance between the root meaning vertex M00001740 with synset \{entity\} and the word vertex $w$ in the graph $M\cup\mathcal{J}\left[R(W),w\right]$.

\subsubsection{Polysemy}

The polysemy measures the number of different meanings possessed by
a given word. About 44\% of English words are polysemous, which means
that they have more than one meaning \cite{Britton1978}. The log~transformed value of polysemy quantifies the missing bits of information
required for correct understanding of the intended meaning of a given
word. That missing information is usually extracted from the context
of the conversation. For monosemous words, which have only one meaning,
there is no ambiguity and no information is missing.
In graph theoretic notation, the polysemy is
\begin{equation}
\text{Polysemy}(w)=\left|\mathcal{A}(W,w)\right|
\end{equation}
which counts the number of all meaning vertices that are adjacent to the
word vertex $w$.

\subsubsection{Information content}

The \emph{intrinsic} information content of words or meanings is measured in bits solely from the graph-theoretic structure of WordNet~3.1.
For comparison of different formulas that measure the information content of words in WordNet~3.1, however, it is helpful to work with normalized values in the unit interval $[0,1]$.
In order to write compactly the formulas for information content, we will need the following word functions.

For a given word~$w$: $\mathcal{S}\left(w\right)$ lists the meaning
subsumers in the graph $M\cup\mathcal{J}\left[R(W),w\right]$, $\mathcal{V}\left(w\right)$ lists the meaning subvertices
in the graph $M\cup\mathcal{J}(W,w)$, $\mathcal{H}\left(w\right)$
lists the meaning hyponyms in the set difference $\mathcal{V}\left(w\right)\backslash\left[\mathcal{A}(W,w)\cup w\right]$,
$\mathcal{L}\left(w\right)$ lists the leaves in the graph $M\cup\mathcal{J}(W,w)$, and $\mathcal{C}(w)$ computes the commonness
$\sum_{i\in\mathcal{L}(G,w)}\frac{1}{\mathcal{S}(M,i)}$ in
the graph $G=M\cup\mathcal{J}(W,w)$.

Several constants specific for WordNet~3.1 are also useful: $\mathcal{V}_{\max}=82192$
is the maximal number of meaning subvertices, $\mathcal{L}_{\max}=65031$
is the maximal number of leaves, $\mathcal{C}_{\min}=1/35$ is the
minimal commonness, and $\mathcal{C}_{\max}=6863.6$ is the maximal
commonness.

Below, we summarize seven information content formulas whose performance
for the analysis of creativity in design review conversations has been tested previously
\cite{Gogo2018}.

Information content by Blanchard et al. \cite{Blanchard2008}:
\begin{equation}
IC(w)=1-\frac{\log\left|\mathcal{L}(w)\right|}{\log\left(\mathcal{L}_{\max}\right)}
\end{equation}
 
Information content by Meng et al. \cite{Meng2012}:
\begin{equation}
IC(w)=\frac{\log\left[\mathcal{T}(w)\right]}{\log\left(\mathcal{T}_{\max}\right)}\left[1-\frac{\log\left[1+\sum _{i\in\mathcal{H}(w)}\frac{1}{\mathcal{T}(i)}\right]}{\log\left(\mathcal{V}_{\max}\right)}\right]
\end{equation}
 
Information content by S\'{a}nchez et al. \cite{Sanchez2011}:
\begin{equation}
IC(w)=\frac{\log\left(\mathcal{L}_{\max}\right)+\log\left|\mathcal{S}(w)\right|-\log\left|\mathcal{L}(w)\right|}{\log\left(\mathcal{L}_{\max}\right)-\log\left(\mathcal{C}_{\min}\right)}
\end{equation}
 
Information content by S\'{a}nchez and Batet \cite{Sanchez2012}:
\begin{equation}
IC(w)=\frac{\log\left(\mathcal{C}_{\max}\right)-\log\left[\mathcal{C}(w)\right]}{\log\left(\mathcal{C}_{\max}\right)-\log\left(\mathcal{C}_{\min}\right)}
\label{eq:6}
\end{equation}
 
Information content by Seco et al. \cite{Seco2004}:
\begin{equation}
IC(w)=1-\frac{\log\left|\mathcal{V}(w)\right|}{\log\left(\mathcal{V}_{\max}\right)}
\end{equation}
 
Information content by Yuan et al. \cite{Yuan2013}:
\begin{equation}
IC(w)=\frac{\log\left[\mathcal{T}(w)\right]}{\log\left(\mathcal{T}_{\max}\right)}\left(1-\frac{\log\left|\mathcal{L}(w)\right|}{\log\left(\mathcal{L}_{\max}\right)}\right)+\frac{\log\left|\mathcal{S}(w)\right|}{\log\left(\mathcal{V}_{\max}\right)}
\end{equation}

Information content by Zhou et al. \cite{Zhou2008a}:
\begin{equation}
IC(w)=\frac{1}{2}\left[1-\frac{\log\left|\mathcal{V}(w)\right|}{\log\left(\mathcal{V}_{\max}\right)}+\frac{\log\left[\mathcal{T}(w)\right]}{\log\left(\mathcal{T}_{\max}\right)}\right]
\end{equation}

\subsection{Semantic similarity for word pairs}

For a semantic network composed of $n$ word vertices, the average
semantic similarity for all pairs of vertices could be determined
with $(n^{2}-n)/2$ searches in WordNet. Different definitions of
semantic similarity between a pair of distinct words $w_{1}$ and
$w_{2}$ rely on the lowest common subsumer of $w_{1}$ and $w_{2}$,
the fraction of common subsumers, or the shortest path distance between
$w_{1}$ and $w_{2}$ \cite{Gogo2018}.

The lowest common subsumer $\mathcal{K}\left(w_{1},w_{2}\right)$
of a pair of distinct words $w_{1}$ and $w_{2}$ in the graph $G=M\cup\mathcal{J}\left[R(W),\left\{ w_{1},w_{2}\right\} \right]$
is the deepest meaning vertex in the taxonomy among all vertices $i$
whose sum $\mathcal{D}\left[G,i,w_{1}\right]+\mathcal{D}\left[G,i,w_{2}\right]$ is minimal. If several common subsumers of $w_{1}$ and $w_{2}$ are
at the same depth in the WordNet~3.1 taxonomy, the meaning vertex
with lowest entry number is considered to be the unique $\mathcal{K}(w_{1},w_{2})$. The depth $\mathcal{T}\left[\mathcal{K}\left(w_{1},w_{2}\right)\right]$ of the lowest common subsumer $\mathcal{K}\left(w_{1},w_{2}\right)$ is determined solely within the subgraph $M$.

The shortest path distance $\mathcal{D}\left(w_{1},w_{2}\right)$
between a pair of distinct words $w_{1}$ and $w_{2}$ in the graph
$U(M)\cup\mathcal{J}\left[U(W),\left\{ w_{1},w_{2}\right\} \right]$
is the number of edges on the shortest path with subtracted two-edge
contribution outside the subgraph $U(M)$.

\subsubsection{Path-based similarity measures}

Semantic similarity by Al-Mubaid and Nguyen \cite{AlMubaid2006}:
\begin{equation}
\text{sim}\left(w_{1},w_{2}\right)=1-\frac{\log\left[1+\mathcal{D}\left(w_{1},w_{2}\right)\left\{ \mathcal{T}_{\max}-\mathcal{T}\left[\mathcal{K}(w_{1},w_{2})\right]\right\} \right]}{\log\left[1+2\left(\mathcal{T}_{\max}-1\right)^{2}\right]}
\end{equation}
Semantic similarity by Leacock and Chodorow \cite{Leacock1998}:
\begin{equation}
\text{sim}\left(w_{1},w_{2}\right)=1-\frac{\log\left[\mathcal{D}(w_{1},w_{2})+1\right]}{\log\left[2\mathcal{T}_{\max}-1\right]}
\end{equation}
Semantic similarity by Li et al. \cite{Li2003}:
\begin{equation}
\text{sim}\left(w_{1},w_{2}\right)=
e^{-0.2\mathcal{D}(w_{1},w_{2})}
\frac{e^{1.2\mathcal{T}\left[\mathcal{K}(w_{1},w_{2})\right]} -1}{e^{1.2\mathcal{T}\left[\mathcal{K}(w_{1},w_{2})\right]} +1}
\end{equation}
Semantic similarity by Rada et al. \cite{Rada1989}:
\begin{equation}
\text{sim}\left(w_{1},w_{2}\right)=1-\frac{\mathcal{D}\left(w_{1},w_{2}\right)}{2\left(\mathcal{T}_{\max}-1\right)}
\end{equation}
Semantic similarity by Wu and Palmer \cite{Wu1994}:
\begin{equation}
\text{sim}\left(w_{1},w_{2}\right)=\frac{2\left\{ \mathcal{T}\left[\mathcal{K}(w_{1},w_{2})\right]-1\right\} }{2\left\{ \mathcal{T}\left[\mathcal{K}(w_{1},w_{2})\right]-1\right\} +\mathcal{D}(w_{1},w_{2})}
\end{equation}

\subsubsection{Subsumer-based similarity measures}

Subsumer-based similarity measures reduce to path-based ones
only for monosemous words, whereas for polysemous words they produce distinct results.

Semantic similarity by Jaccard \cite{Jaccard1912}:
\begin{equation}
\text{sim}\left(w_{1},w_{2}\right)=\frac{\left|\mathcal{S}\left(w_{1}\right)\cap\mathcal{S}\left(w_{2}\right)\right|}{\left|\mathcal{S}\left(w_{1}\right)\cup\mathcal{S}\left(w_{2}\right)\right|}
\end{equation}

Semantic similarity by Braun-Blanquet \cite{BraunBlanquet1932}:
\begin{equation}
\text{sim}\left(w_{1},w_{2}\right)=\frac{\left|\mathcal{S}\left(w_{1}\right)\cap\mathcal{S}\left(w_{2}\right)\right|}{\max\left[\left|\mathcal{S}\left(w_{1}\right)\right|,\left|\mathcal{S}\left(w_{2}\right)\right|\right]}
\end{equation}
 
Semantic similarity by Dice \cite{Dice1945}:
\begin{equation}
\text{sim}\left(w_{1},w_{2}\right)=\frac{2\left|\mathcal{S}\left(w_{1}\right)\cap\mathcal{S}\left(w_{2}\right)\right|}{\left|\mathcal{S}\left(w_{1}\right)\right|+\left|\mathcal{S}\left(w_{2}\right)\right|}
\end{equation}
 
Semantic similarity by Otsuka and Ochiai \cite{Otsuka1936,Ochiai1957}:
\begin{equation}
\text{sim}\left(w_{1},w_{2}\right)=\frac{\left|\mathcal{S}\left(w_{1}\right)\cap\mathcal{S}\left(w_{2}\right)\right|}{\sqrt{\left|\mathcal{S}\left(w_{1}\right)\right|\left|\mathcal{S}\left(w_{2}\right)\right|}}
\end{equation}
 
Semantic similarity by Kulczy\'{n}ski \cite{Kulczynski1927}:
\begin{equation}
\text{sim}\left(w_{1},w_{2}\right)=\frac{\left|\mathcal{S}\left(w_{1}\right)\cap\mathcal{S}\left(w_{2}\right)\right|}{2}\left(\frac{1}{\left|\mathcal{S}\left(w_{1}\right)\right|}+\frac{1}{\left|\mathcal{S}\left(w_{2}\right)\right|}\right)
\end{equation}
 
Semantic similarity by Simpson \cite{Simpson1960}:
\begin{equation}
\text{sim}\left(w_{1},w_{2}\right)=\frac{\left|\mathcal{S}\left(w_{1}\right)\cap\mathcal{S}\left(w_{2}\right)\right|}{\min\left[\left|\mathcal{S}\left(w_{1}\right)\right|,\left|\mathcal{S}\left(w_{2}\right)\right|\right]}
\end{equation}
 
\subsubsection{Information content-based similarity measures}

The following five information content-based similarity formulas could
take as an input any of the seven information content formulas to
give a total of 35~different information content-based similarity
measures whose performance for the analysis of creativity in design review conversations
has been tested previously \cite{Gogo2018}.

Semantic similarity by Jiang and Conrath \cite{Jiang1997}:
\begin{equation}
\text{sim}\left(w_{1},w_{2}\right)=1 -\frac{1}{2}\left\{ IC\left(w_{1}\right)+IC\left(w_{2}\right)-2IC\left[\mathcal{K}\left(w_{1},w_{2}\right)\right]\right\} 
\end{equation}
 
Semantic similarity by Lin \cite{Lin1998}:
\begin{equation}
\text{sim}\left(w_{1},w_{2}\right)=\frac{2IC\left[\mathcal{K}\left(w_{1},w_{2}\right)\right]}{IC\left(w_{1}\right)+IC\left(w_{2}\right)}
\label{eq:22}
\end{equation}
 
Semantic similarity by Meng et al. \cite{Meng2014}:
\begin{equation}
\text{sim}\left(w_{1},w_{2}\right)=\left\{ \frac{2IC\left[\mathcal{K}\left(w_{1},w_{2}\right)\right]}{IC\left(w_{1}\right)+IC\left(w_{2}\right)}\right\} ^{\frac{1-\exp\left[-0.08\,\mathcal{D}\left(w_{1},w_{2}\right)\right]}{\exp\left[-0.08\,\mathcal{D}\left(w_{1},w_{2}\right)\right]}}
\end{equation}
 
Semantic similarity by Resnik \cite{Resnik1995}:
\begin{equation}
\text{sim}\left(w_{1},w_{2}\right)=IC\left[\mathcal{K}\left(w_{1},w_{2}\right)\right]
\label{eq:24}
\end{equation}
 
Semantic similarity by Zhou et al. \cite{Zhou2008b} is a weighted average of $k\times$~path-based Leacock--Chodorow similarity and $(1-k)\times$ information content-based Jiang--Conrath similarity.
Usually, the two weights are set to be equal $k=1-k=\frac{1}{2}$.

\pagebreak
\section{Correlation between subjective and objective evaluation of word similarity}

\subsection{Cluster analysis based on RG-65 dataset}

\begin{figure}[t!]
\begin{centering}
\includegraphics[width=\textwidth]{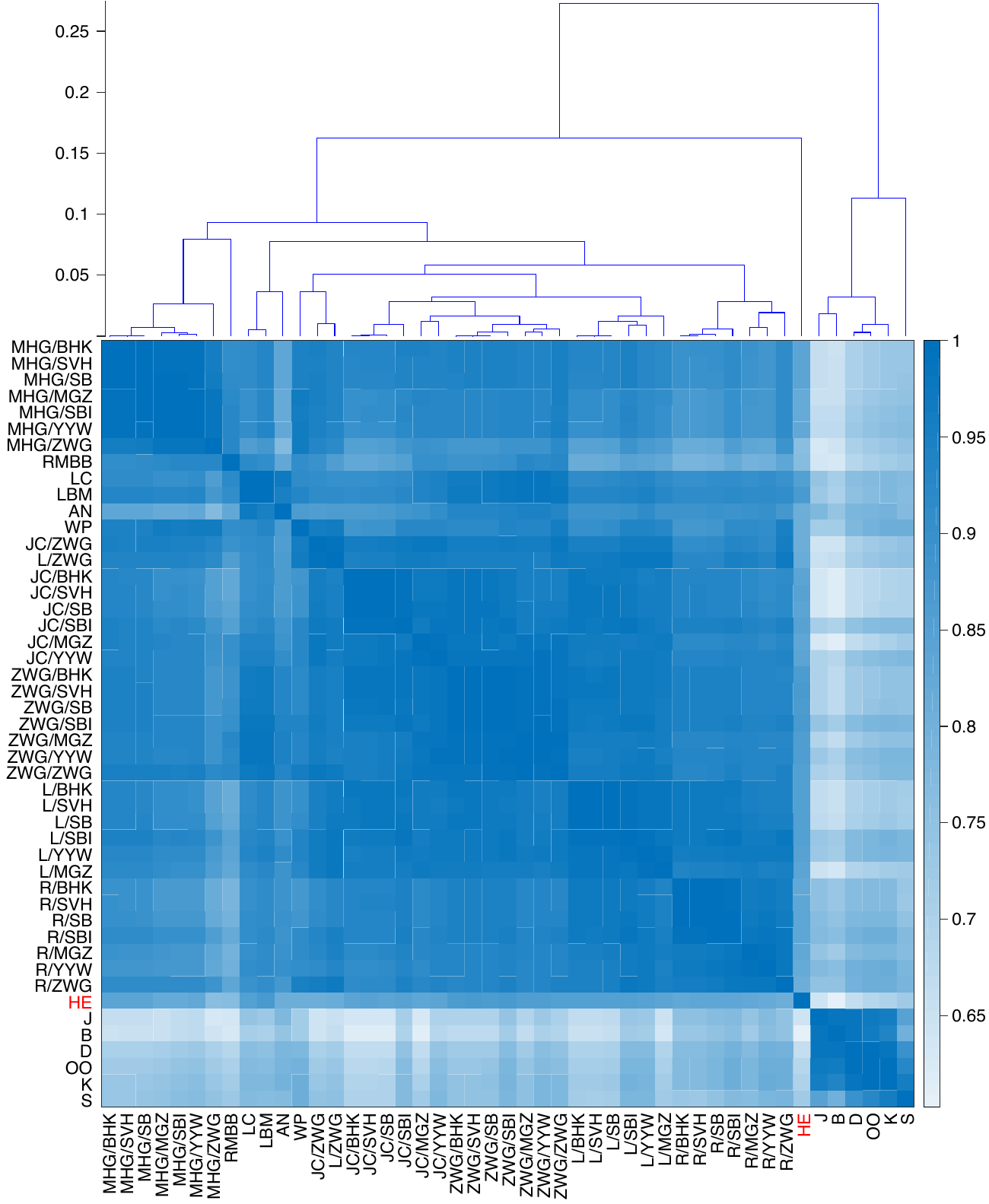}
\par\end{centering}

\caption{\label{fig:5}Pearson correlation matrix map with hierarchical clustering (dendrogram) based on the Pearson correlation distance between subjective human evaluation (HE) of word similarity for noun--noun pairs in the \mbox{RG-65} dataset and 46~objective semantic similarity measures computed with the use of WordNet~3.1. The similarity measures segregate into two clusters, a larger cluster composed from information content-based or path-based similarity measures, and a smaller cluster composed from subsumer-based similarity measures. Formulas for the similarity measures are provided in the main text.
Abbreviations:
AN:~Al-Mubaid--Nguyen,
B:~Braun-Blanquet,
BHK:~Blanchard--Harzallah--Kuntz,
D:~Dice,
J:~Jaccard,
JC:~Jiang--Conrath,
K:~Kulczy\'{n}ski,
L:~Lin,
LBM:~Li--Bandar--McLean,
LC:~Leacock--Chodorow,
MGZ:~Meng--Gu--Zhou,
MHG:~Meng--Huang--Gu,
OO:~Otsuka--Ochiai,
R:~Resnik,
RMBB:~Rada--Mili--Bicknell--Blettner,
S:~Simpson,
SB:~S\'{a}nchez--Batet,
SBI:~S\'{a}nchez--Batet--Isern,
SVH:~Seco--Veale--Hayes,
WP:~Wu--Palmer,
YYW:~Yuan--Yu--Wang,
ZWG:~Zhou--Wang--Gu.}
\end{figure}

The development of many semantic similarity measures based on the hypernym-hyponym hierarchy in WordNet was done by their authors using as a testbed the \mbox{RG-65} dataset containing 65 noun--noun pairs, whose semantic similarity is evaluated from subjective reports collected from human subjects \cite{Rubenstein1965}.
Consequently, we have also used the RG-65 dataset to assess the degree of correlation between subjective human evaluation of word
similarity and all semantic similarity measures reported in this work.
Pearson correlation analysis shows that subsumer-based similarity measures are least correlated with subjective human evaluation (average $r=0.67$, $P<0.001$), followed by path-based similarity measures (average $r=0.82$, $P<0.001$), and information content-based similarity measures (average $r=0.84$, $P<0.001$).
Consequent hierarchical clustering segregates subsumer-based
similarity measures into a single small cluster that is least correlated
to human evaluation, but mixes path-based and information content-based similarity measures into another large cluster (Fig.~\ref{fig:5}).
Although any of the available path-based or information content-based
similarity measures could be used for exploration of divergent or
convergent thinking in conversational transcripts, previous experimental
study of creative cognition by design students in real-world educational
setting has found that information content-based similarity
measures exhibit highest statistical power to differentiate between
successful and unsuccessful ideas \cite{Gogo2018}.
{Here, we have identified the information content formula by S\'{a}nchez--Batet~\eqref{eq:6}, as the one that exhibits highest correlation with human evaluation of word similarlity, with an average $r=0.85$ across all five information content-based semantic similarity formulas. Furthermore, the semantic similarity formula by Lin~\eqref{eq:22} has the highest $r=0.85$ when used with S\'{a}nchez--Batet formula~\eqref{eq:6} among all purely information content-based semantic similarity formulas, which are computationally fast to execute in real-time application. Thus, the combination of formulas \eqref{eq:6} and \eqref{eq:22} ensures best correlation with human evaluation of similarlity and optimizes computational speed for engineering applications.}

\section{Aim 3: Back-testing dynamic semantic measures for success of design ideas}

\subsection{Construction of semantic networks from design conversations}

The concurrent verbalization may not access all cognitive processes involved in design thinking.
Nevertheless, the language provides an output channel of information that could be used to monitor the ongoing design process and an input channel of information that could be used by the designer to incorporate external feedback on the designed product.
Thus, it is desirable to assess whether verbalization and language could be useful in aiding the design process, even though they do not exhaust everything that goes on in the designer's mind.
Next, we illustrate with a concrete empirical example how design review conversations could be analyzed using moving time window and demonstrate the utility of dynamic semantic measures to differentiate between successful and unsuccessful ideas.

\subsection{Distinguishing successful ideas from unsuccessful ideas}

\begin{figure}[t!]
\begin{centering}
\includegraphics[width=\textwidth]{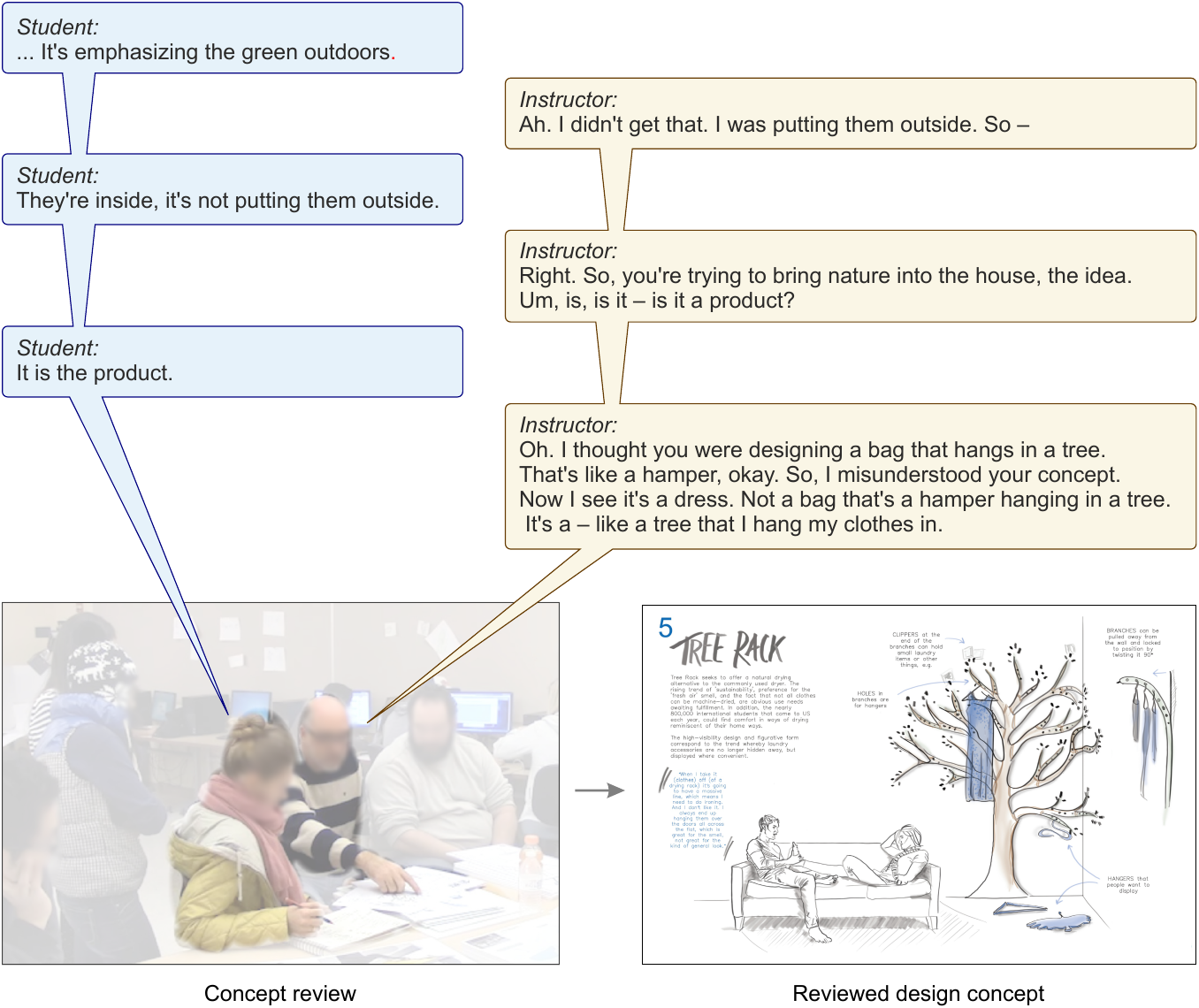}
\par\end{centering}
\caption{\label{fig:6} Part of design conversation (concept review) from DTRS~10 dataset.}
\end{figure}

For numerical analysis, we have employed the experimental dataset of complete transcripts with design review conversations
provided as a part of the 10th Design Thinking Research Symposium (DTRS~10) including two subsets with students majoring in Industrial Design: a subset with 7 junior students and a subset with 5 graduate students \cite{Adams2013,Adams2015}.
Each design project consisted of five stages: (1) task review, (2) concept review, (3) client review, (4) concept reduction review, and (5) final presentation. For each project, the students developed several possible design solutions (Fig.~\ref{fig:6}), from which only the best one was selected to appear in the final presentation.

This final design solution, which was selected after consultation with the client, is considered to be the \emph{successful idea} because it has won the competition with other design solutions (ideas) that were not successful in regard to appearing in the final presentation. Here, our main motivation is to distinguish the best design solution from all the rest.
Segmentation of the transcripts with regard to successful and unsuccessful ideas was performed based on the videos and the presentation slides.
Overall, there were 12~successful ideas, and 41~unsuccessful ideas in the DTRS~10 dataset \cite{Gogo2018}.

\subsection{Construction of moving time window}

In order to test for possible relationship between attributes of divergent/convergent thinking and the success of design ideas,
we have split the design conversation transcripts for each idea and have constructed dynamic semantic networks with a moving time window that contains 6~distinct nouns.
This approach is different from bag-of-words, because it does not keep multiplicity of nouns. The first time window is constructed by removing repeated nouns in the conversation until there are collected 6~distinct nouns.
Average (mean) information content of the 6~nouns in each time window was computed using the S\'{a}nchez--Batet formula~\eqref{eq:6}, whereas average (mean) semantic similarity of the 15~noun pairs was computed with the Lin formula~\eqref{eq:22}. These particular formulas have been also found to differentiate well between successful and unsuccessful ideas when the design conversations are split into 3 equal parts \cite{Gogo2018}.

The temporal duration for the development of each idea was normalized within the unit interval, so that time=0 indicates the start and time=1 indicates the end of the design review conversation that pertains to the idea under consideration. The shortest time step corresponds to the appearance of the next noun in the conversation. If the next noun is already repeated in the preceding time window, the time step is added but there is no dynamic change of the semantic network. Alternatively, if the next noun is not repeated in the preceding time window, then both the time step is added and there is a dynamic change of the semantic network.

\begin{figure}[t!]
\begin{centering}
\includegraphics[width=\textwidth]{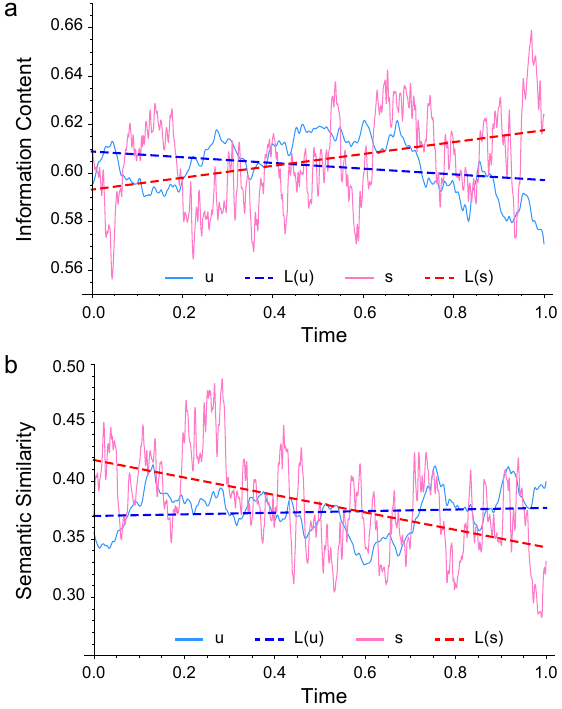}
\par\end{centering}
\caption{\label{fig:7}Dynamics of semantic measures for successful ideas (selected for final presentation) or unsuccessful ideas (not selected for final presentation) computed from design review conversations. (A) The information content increases in time for successful ideas, whereas it decreases for unsuccessful ideas. (B) The semantic similarity decreases in time for successful ideas, whereas it increases for unsuccessful ideas. Legend: s, average trajectory of successful ideas; u, average trajectory of unsuccessful ideas; L(s), linear best fit of successful ideas; L(u), linear best fit of unsuccessful ideas. The averages are based on $n=12$ student projects in Industrial Design from DTRS~10 dataset.}
\end{figure}

Because different students had generated different numbers of unsuccessful ideas, to ensure equal weight of each project we have first averaged the dynamic trajectories per student and only then we have averaged the trajectories of unsuccessful ideas across students.
Despite that the resulting average trajectories of semantic measures appear to be noisy, it is possible to extract smooth trendlines using linear best fit that minimizes the sum of squared residuals (Fig.~\ref{fig:7}).

\subsection{Statistical differences in the rates of change of semantic measures}

To test whether the linear trendlines constructed with moving time window are able to extract faithfully the rates of change (slopes of the trendlines) for information content and semantic similarity reported in previous study where the design review conversations were divided into 3 equal parts \cite{Gogo2018}, we have employed one-tailed paired t-tests. The statistical analysis indeed confirmed that the information content exhibited positive rate of change ($k_s=0.024$) for successful ideas, whereas it exhibited negative rate of change ($k_u=-0.012$) for unsuccessful ideas, ($t=2.24$, $P=0.024$). Opposite dynamics was observed for semantic similarity,
which exhibited negative rate of change ($k_s=-0.08$) for successful ideas and positive rate of change ($k_u=0.01$) for unsuccessful ideas, ($t=-2.05$, $P=0.032$). 
Computationally, divergent thinking is identified by negative rate of change of semantic similarity in time, whereas convergent thinking is identified by positive rate of change of semantic similarity in time.
Thus, consistent with psychological theories linking creativity with divergent thinking \cite{Guilford1957,Hudson1974,Runco2004,Runco2007,Runco2020}, we have found that successful ideas exhibit computational attributes of divergent thinking such as increasing information content (Fig.~\ref{fig:7}a) and decreasing semantic similarity (Fig.~\ref{fig:7}b) during the development of ideas in time, whereas unsuccessful ideas exhibit computational attributes of convergent thinking such as decreasing information content (Fig.~\ref{fig:7}a) and increasing semantic similarity (Fig.~\ref{fig:7}b).
These results were obtained in retrospective fashion through analysis of human curated transcripts, which eliminated machine errors in speech-to-text conversion and also performed post-processing of nouns such as conversion of plural to singular and omission of nouns that are absent from WordNet~3.1.
However, the numerical plots establish as a proof of principle that conversation analysis could be employed in real time and the trendlines for semantic measures could be provided to the designer as a future forecast of whether the design product is going to be successful.

\section{Discussion}

\subsection{WordNet captures faithfully the distinction between words and meanings}

Progress on difficult scientific problems usually requires the development and adoption of new research tools for investigation \cite{Laudan1978,Marx2013,Glocker2021}. Here, our goal was to lay the foundations of an objective methodology for approaching the problem of human creativity. 
While the idea of using WordNet's hypernymy for the investigation of analogical concepts is not new \cite{Geum2016}, 
here we have scrutinized the possible implementation of dynamic semantic networks based on WordNet~3.1 as a tool for the exploration of creative thinking through analysis of verbal data obtained concurrently with the act of problem solving.
The graph theoretic representation of WordNet~3.1 as a composition of two directed subgraphs, respectively for \emph{words} and \emph{meanings}, is computationally powerful and neuroscientifically well-tailored to capture the two anatomically distinct language-related brain cortical areas specialized for functional processing of \emph{words} and \emph{meanings} \cite{Gogo2018,Georgiev2021}. This is to be contrasted with the prevalent natural language processing approaches whose main goal is to analyze the \emph{meanings} extracted from the verbal data, while viewing the \emph{words} only as labels for the intended meanings that need to be disambiguated by a special preprocessing step of the transcribed texts.
By keeping both \emph{words} and \emph{meanings}, the presented approach captures more faithfully the complexity of human thinking and provides an inroad to virtually imaged concepts that were not verbalized but provide links in the WordNet~3.1 intrinsic hierarchy \cite{Yamamoto2009,Gogo2010}.

\subsection{Advantages of dynamic semantic networks}

Having thoroughly discussed the theoretical and applied aspects of dynamic semantic networks,
we could summarize their main advantages as follows:
(1) reproducible objective analysis of verbal data,
(2) extraction of both verbalized and virtually imaged
concepts in creative problem solving, 
(3) minimal confounding injection
of interpretation at the stage of data analysis due to dual use of
words and meanings, 
(4) minimal disturbance of spontaneous creative cognition
due to reliance on verbalization of naturally occurring inner monologue,
(5) explicit acknowledgment of educational status and language proficiency
of test subjects, and 
(6) possibility for real-time computer-assisted
audio-visual feedback of the constructed dynamic semantic network
for enhancement of human creativity. 

\subsection{Limitations of dynamic semantic networks}

The main limitations of dynamic semantic networks include:
(1) lower performance with certain types of creativity that rely on sensual inner stream of consciousness composed of visual images or sounds instead of words (e.g. painting of art or composing of music), 
(2) possibly inadequate testing of individuals with low educational status, low language-proficiency or neurological deficits leading to aphasia, and 
(3) lack of direct access to neural processes that remain outside of the contents of individual conscious experiences \cite{Georgiev2017,Georgiev2020,Georgiev2020b}. 
These cons provide an effective definition of the domain of applicability of semantic networks. For supporting creative cognition, inference methodologies based on semantic distance can be employed \cite{Sarica2021}.
For studying creative cognition outside of the domain of individual conscious experiences, dynamic semantic networks could be complemented with mind-reading technologies that rely on reconstruction of mental images (e.g., visual images) from recorded electrical brain activity \cite{Horikawa2017,Roelfsema2018}.

\subsection{Outlook for future work}

The range of creative activities that could benefit from dynamic analysis with semantic networks is quite extensive and includes much of problem solving in science, technology, engineering and mathematics. Economically most important forms of creativity related to design
and innovation of cutting-edge products, equipment or services,
is performed by highly-trained, well-educated professionals with excellent language proficiency.
Therefore, their creative performance is subject to verbalization,
modeling and improvement with dynamic semantic networks, which substantiates
the need of future wider adoption of semantic networks in theoretical and applied cognitive science.
This can be connected with an interdisciplinary approach to design thinking.
Computer systems endowed with general artificial intelligence may dynamically monitor semantic measures of verbalized inner monologue, e.g. if the designer thinks aloud, and provide real-time feedback on creative problem solving. Human creativity could be then enhanced through suggestions that lead to divergence of semantic similarity of the developed solutions.

\section{Conclusions}

{This work sought to develop a complete workflow for real-time application of dynamic semantic networks for monitoring cognitive processes during creative design, using specific measures on these networks that correlate with human evaluation. To achieve that objective, we back-tested the actual performance of the developed workflow evaluating ideas generated in design review conversations from an established dataset. This testing involved construction of semantic networks from design conversations, distinguishing successful ideas from unsuccessful ones, and construction of moving time window. The results demonstrate statistical differences in the rate of change of semantic measures for successful ideas and unsuccessful ideas. Overall, successful ideas exhibit computational attributes pertaining to divergent thinking, while unsuccessful ideas exhibit attributes of convergent thinking. This is seen as a proof of principle that dynamic analysis of conversations can be employed in real-time as a future forecast of the success of ideas and design products. This workflow allows for objective analysis of verbal data in design while preserving the spontaneity of creative cognition. This opens up the possibility of real-time AI tools that analyze and enhance human creativity as it occurs during the design process.}


\paragraph{Funding Statement}

This research received no external funding.

\paragraph{Competing Interests}

The authors declare none.

\paragraph{Data Availability Statement}

WordNet~3.1 is available online at: \url{https://wordnet.princeton.edu/}

The RG-65 dataset \cite{Rubenstein1965} for human judgements of word similarity is available online at: \url{https://doi.org/10.1145/365628.365657}

The authors have signed Data-Use Agreements to Dr. Robin Adams (Purdue University) for accessing the Purdue DTRS Design Review Conversations Database, thereby agreeing not to reveal personal identifiers and not to create any commercial products.




\end{document}